\documentclass[journal,twoside,web]{ieeecolor}
\usepackage{generic}
\usepackage{cite}
\usepackage{amsmath,amssymb,amsfonts}
\usepackage{algorithm}
\usepackage{algorithmic}
\usepackage{graphicx}
\usepackage{hyperref}
\hypersetup{hidelinks}
\usepackage{textcomp}
\usepackage{multirow}
\usepackage{tabularx}
\usepackage{array}

\providecommand{\refname}{References}

\def\BibTeX{{\rm B\kern-.05em{\sc i\kern-.025em b}\kern-.08em
    T\kern-.1667em\lower.7ex\hbox{E}\kern-.125emX}}
\begin{document}
\title{SGAM: Shared Gaussian Geometry with Implicit Amplitude Modeling for Scan-Specific 3D Multi-Contrast MRI Reconstruction}
\author{Jingran Xu, Dong Liang, \IEEEmembership{Senior Member, IEEE}, Hairong Zheng, \IEEEmembership{Senior Member, IEEE}, Yuanyuan Liu, and Yanjie Zhu, \IEEEmembership{Senior Member, IEEE}
\thanks{This study was supported in part by the National Key R\&D Program of China under Grant No. 2023YFA1011403; in part by the National Natural Science Foundation of China under Grant Nos. 62322119, 62531024, and 12226008; in part by the Guangdong Basic and Applied Basic Research Foundation under Grant No. 2025A1515012966. (Corresponding author: Yuanyuan Liu, Yanjie Zhu)} 
\thanks{Jingran Xu, Yuanyuan Liu, and Yanjie Zhu are with Paul C. Lauterbur Research Center for Biomedical Imaging, Shenzhen Institutes of Advanced Technology, Chinese Academy of Sciences, Shenzhen, Guangdong, China (e-mail: {jr.xu; liuyy; yj.zhu}@siat.ac.cn).}
\thanks{Dong Liang is with the Research Center for Medical AI, Shenzhen Institutes
of Advanced Technology, Chinese Academy of Sciences, Shenzhen, China (e-mail: dong.liang@siat.ac.cn).}
\thanks{Hairong Zheng is with Nanjing University, Nanjing, Jiangsu, China (email: Hr.zheng@nju.edu.cn).}
}

\maketitle

\begin{abstract}
Three-dimensional (3D) multi-contrast magnetic resonance imaging (MCMRI) provides rich anatomical and quantitative information but requires long acquisition times, motivating k-space undersampling. However, reconstruction of large volumetric datasets imposes substantial computational and memory demands. To address this challenge, we propose SGAM, a memory-efficient, scan-specific framework for joint full-volume 3D MCMRI reconstruction. SGAM is based on a shared-geometry Gaussian representation in which amplitudes are modeled by a multi-output implicit neural representation (INR) and contrast-specific phases by explicit variables. This design exploits common anatomical structure across contrasts while preserving contrast-specific signal variations. The representation is jointly optimized using only the acquired multi-coil k-space without external training data. Experiments showed that SGAM consistently outperformed the comparison methods across imaging tasks and acceleration factors, with greater improvements under stronger undersampling. SGAM also achieved a favorable balance between reconstruction quality and computational cost, demonstrating its effectiveness for 3D multi-contrast MRI reconstruction. 
\end{abstract}


\section{Introduction}
\label{sec:introduction}
\IEEEPARstart{T}{hree}-dimensional (3D) multi-contrast magnetic resonance imaging (MCMRI) is an important imaging technique that enables applications such as multiparametric quantitative mapping~\cite{ye2022multi,zhang2024mclaro}, and water-fat separation~\cite{li2022accelerating}. However, the acquisition of multiple 3D contrasts is time-consuming, which greatly hinders its clinical use~\cite{guo2023joint,yang2022fast}. Undersampling k-space is widely used to accelerate 3D MCMRI~\cite{ccukur2025tutorial}. Nevertheless, the large volumetric data size and severe information loss under high acceleration make accurate recovery increasingly challenging. Thus, efficient reconstruction from highly undersampled multi-contrast k-space data is highly desirable.

Conventional iterative reconstruction methods typically employ handcrafted priors, such as image sparsity and low-rankness across different contrasts in MCMRI, to regularize the reconstruction~\cite{lustig2007sparse,lingala2011accelerated,tamir2017t2}. However, their performance often degrades at high acceleration factors, and the design of effective priors is nontrivial. Recent advances in deep learning-based MRI reconstruction provide a promising alternative by learning image priors directly from data, enabling improved reconstruction under more aggressive undersampling. Among them, supervised learning methods can learn strong image priors from large-scale training datasets~\cite{hammernik2018learning, aggarwal2018modl,knoll2020deep, polak2020joint}, but for 3D MCMRI, collecting fully sampled training data is challenging and time-consuming~\cite{niessen2025inr}. Self-supervised and scan-specific methods have been developed to learn directly from undersampled measurements or individual scans, thereby reducing the dependence on fully sampled training data~\cite{yaman2022zero,darestani2021accelerated}. For example, self-supervision via data undersampling (SSDU) partitions the acquired k-space measurements into two disjoint subsets, one for data consistency and the other for loss calculation, enabling network training using only undersampled data~\cite{yaman2020self}. IMJENSE introduced a scan-specific implicit neural representation (INR) framework that represents MR images using coordinate-based neural networks and optimizes the network parameters directly from the acquired k-space measurements, without external training data~\cite{feng2023imjense}. However, when extending these approaches to 3D MCMRI reconstruction, the large volumetric data size can easily exceed GPU memory capacity and requires substantial computational cost for optimization.

Several strategies have been explored to address this issue. A straightforward way is to split the 3D k-space data into 2D slices by applying an inverse Fourier transform along the fully sampled readout direction and reconstruct the resulting slices independently~\cite{niessen2025inr}. This strategy reduces the 3D reconstruction to a series of 2D problems with substantially lower memory and computational demands. However, such slice-wise reconstruction ignores through-plane correlations and may lead to inter-slice discontinuity in the reconstructed volume. Another strategy is using low-dimensional representations such as subspace modeling or tensor decomposition, such that the network reconstructs a small number of subspace coefficients rather than the entire high-dimensional image volumes~\cite{luo2023low}. For example, Zero-DeepSub integrates scan-specific deep reconstruction with low-rank subspace modeling for a 3D-QALAS, with reduced memory and computational cost~\cite{jun2024zero}. Zhang et al. proposed low-rank integrated implicit neural representation (LoREIN), which uses INR to parameterize the spatial bases of a low-rank representation and jointly optimizes image and quantitative-parameter domains~\cite{zhang2026unsupervised}. TenF-INR uses INRs to parameterize the factor functions of a low-rank tensor decomposition, reducing the parameter space and computational burden while preserving multidimensional correlations~\cite{liu2026unsupervised}. However, low-dimensional subspace or tensor models may have limited capacity to capture complex contrast-dependent variations. A compact yet flexible volumetric representation is therefore still needed.

Recently, Gaussian representation has emerged as an efficient explicit representation of volumetric data~\cite{kerbl3Dgaussians}. It models a volume using a set of anisotropic Gaussian primitives with learnable locations, shapes, and features, providing a continuous representation without using the dense voxel-wise grid. Its compact and differentiable formulation enables efficient optimization, making it attractive for high-dimensional MR reconstruction. Peng et al. proposed 3DGSMR, which represents a 3D MR volume using Gaussian primitives and optimizes them with the undersampled k-space measurements in a scan-specific manner~\cite{peng2025three}. Gaussian representation has also been applied for slice-to-volume MRI reconstruction~\cite{dannecker2026fast} and time-resolved volumetric MRI~\cite{xie2026spatiotemporal}, suggesting its potential as an efficient representation for high-dimensional MR reconstruction.

For 3D MCMRI reconstruction, images with different contrasts exhibit contrast-dependent intensity variations while sharing common anatomical structures. Representing each contrast independently ignores such correlations and requires repeated modeling of the same underlying geometry. Therefore, in this study, we propose \textbf{S}hared \textbf{G}aussian geometry with implicit \textbf{A}mplitude \textbf{M}odeling (SGAM), a memory-efficient scan-specific framework for accelerated 3D MCMRI reconstruction. SGAM represents the 3D MCMRI using shared anisotropic Gaussian geometry, implicitly predicted contrast-specific amplitudes, and explicitly optimized contrast-specific phases. The shared geometry captures common anatomical structures and reduces redundant geometric parameterization across contrasts. A shared multi-output amplitude INR jointly models amplitudes across spatial locations and contrasts, while the explicit phase variables preserve contrast-specific phase variations. All parameters are jointly optimized directly from the acquired multi-coil k-space without external training data.

The main contributions of this work are summarized as follows:
\begin{enumerate}
    \item We introduce a shared-geometry Gaussian representation for joint reconstruction of 3D MCMRI, which exploits common anatomical structure across contrasts while preserving contrast-specific signal variations. This enables scan-specific reconstruction without external training data or slice-wise decomposition.
    \item We develop a memory-efficient full-volume 3D reconstruction framework that avoids the excessive GPU memory demands of dense volumetric neural networks while maintaining computational efficiency.
    \item The proposed method consistently improves reconstruction quality at high accelerations across two 3D MCMRI applications, with acceleration factors up to 16$\times$. It also better preserves inter-contrast signal relationships, leading to improved accuracy in the quantitative maps.
\end{enumerate}


\section{Background}
\subsection{Reconstruction Model for Multi-Contrast MRI}
Let $X=[x_1,\ldots,x_E]$ denote a set of 3D multi-contrast MR images, where $x_e$ represents the image of the $e$-th contrast and $E$ is the number of contrasts. For receiver coil $c=1,\ldots,C$, the undersampled k-space measurement $y_{c,e}$ is modeled as~\cite{pruessmann1999sense}
\begin{equation}
 y_{c,e}=M_e\mathcal{F} S_c x_e+\epsilon_{c,e},
 \label{eq:direct_forward}
\end{equation}
where $M_e$ is the binary sampling mask, $\mathcal{F}$ represents the 3D discrete Fourier transform, $S_c$ is the coil sensitivity map, and $\epsilon_{c,e}$ is the noise term. For simplicity, $\mathcal{A}_{c,e}=M_e\mathcal{F} S_c$ denotes the overall forward encoding operation.
 
The reconstruction of $X$ can then be generally formulated as:
\begin{equation}
 \widehat{X}
 = \underset{X}{\operatorname{arg\,min}}\;
 \sum_{e=1}^{E}\sum_{c=1}^{C}
 \left\|\mathcal{A}_{c,e}x_e-y_{c,e}\right\|_2^2
 + \lambda\mathcal{R}(X),
 \label{eq:inverse_problem}
\end{equation}
where the first term enforces consistency with the acquired measurements, $\mathcal{R}(X)$ represents the regularization, and $\lambda$ controls the regularization strength. For multi-contrast MRI, effective reconstruction should exploit the shared anatomy across contrasts while preserving their contrast-specific complex signal variations.

\subsection{3D Gaussian Representation for 3D MRI}
A 3D MRI volume can be represented by the combination of anisotropic Gaussian primitives~\cite{peng2025three}, which is expressed as 
\begin{equation}
 \widehat{x}_e(\mathbf r)
 =\sum_{n=1}^{N} G_{n}(\mathbf r |o_{n},\boldsymbol{\mu}_n, \boldsymbol{\Sigma}_n),
 \label{eq:gaussian_representation}
\end{equation}
where $\mathbf r\in\Omega$ is an image-space location and $N$ is the number of Gaussian primitives. Each Gaussian primitive $G_{n}$ is formulated as
\begin{equation}
G_{n}(\mathbf r | o_{n},\boldsymbol{\mu}_n, \boldsymbol{\Sigma}_n)= 
o_{n} \cdot \exp(-\frac{1}{2} (\mathbf r-\boldsymbol{\mu}_n)^{\mathsf T} \boldsymbol{\Sigma}_n^{-1}(\mathbf r-\boldsymbol{\mu}_n)),
\end{equation}
where $o_{n} \in \mathbb{C}$, $\boldsymbol{\mu}_n\in\mathbb{R}^3$, and $\boldsymbol{\Sigma}_n \in\mathbb{R}^{3\times3}$ are learnable parameters representing the central intensity value, position and covariance matrix, respectively. $o_{n}$ is in the complex domain to match the complex-valued form of MRI. $\boldsymbol{\mu}_n$ defines the position of the Gaussian primitive, while $\boldsymbol{\Sigma}_n$ defines the scale and orientation. $\boldsymbol{\Sigma}_n$ can further be decomposed into scale $\boldsymbol{\sigma}_n$ and rotation $\boldsymbol{R}_n$ as
\begin{equation}
 \boldsymbol{\Sigma}_n=\boldsymbol{R}_n\operatorname{diag}
 (\sigma_{n,1}^2,\sigma_{n,2}^2,\sigma_{n,3}^2)\boldsymbol{R}_n^{\mathsf T},
\end{equation}
where $\boldsymbol{R}_n$ is the rotation matrix, constructed using trainable quaternions. Compared with voxel-based representations, 3D Gaussian representations provide an explicit and compact description of volumetric images using localized anisotropic primitives, and can be efficiently optimized through differentiable rasterization.

\section{Methods}
\label{sec:direct_methods}
\begin{figure*}[t]
\centering
\includegraphics[width=\textwidth]{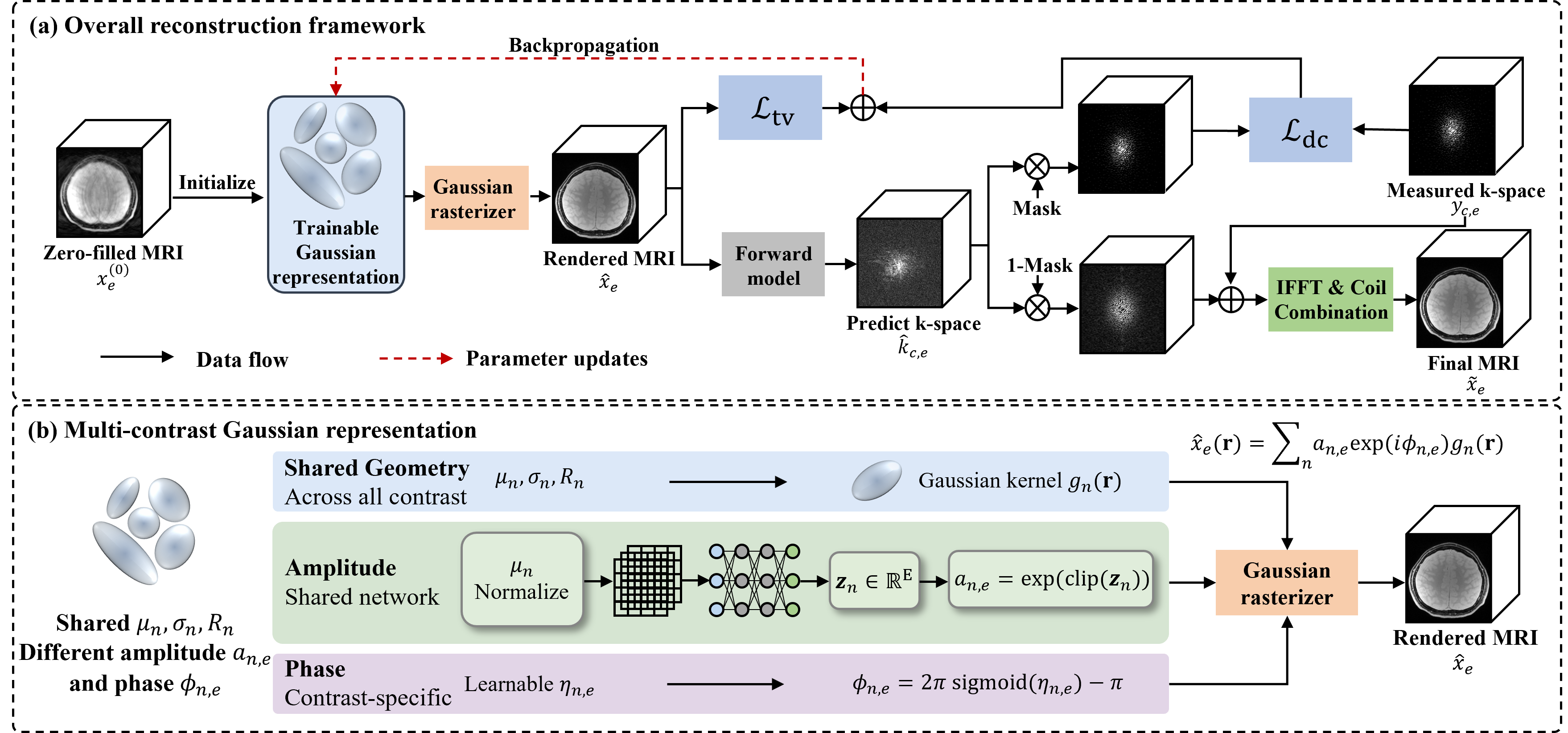}
\caption{Overview of SGAM. (a) Overall reconstruction framework, in which the Gaussian representation is jointly optimized using multi-coil k-space data consistency and magnitude TV regularization, followed by acquired-sample replacement and coil combination. (b) Multi-contrast Gaussian representation comprising shared geometry, implicitly predicted contrast-specific amplitudes, and explicitly optimized phases.}
\label{fig:joint_reconstruction}
\end{figure*}

SGAM is a memory-efficient scan-specific framework for 3D multi-contrast MR reconstruction based on Gaussian representation. The overview of the framework is shown in Fig.~\ref{fig:joint_reconstruction}. First, SGAM represents the common anatomical structure across contrasts using shared 3D Gaussian geometry. The Gaussian centers, scales, and rotations are shared to jointly exploit structural information across contrasts and reduce redundant geometric parameterization. Second, contrast-dependent complex appearance is modeled using implicitly predicted amplitudes and explicitly optimized phases. A shared multi-output amplitude INR models amplitudes across spatial locations and contrasts, while contrast-specific phase variables preserve flexible phase variations. Finally, all parameters are jointly optimized using only the acquired k-space measurements, without external training data.

\subsection{Shared Gaussian Geometry for Multi-Contrast MRI}
For the $e$-th contrast, SGAM renders the 3D MRI as
\begin{equation}
 \widehat{x}_e(\mathbf r)
 =\sum_{n=1}^{N} a_{n,e}\exp(\mathrm{i}\phi_{n,e})g_n(\mathbf r),
 \label{eq:direct_image}
\end{equation}
where $g_n(\mathbf{r})$ is an anisotropic Gaussian kernel defined by its center $\boldsymbol{\mu}_n$, scale $\boldsymbol{\sigma}_n$, and rotation $\boldsymbol{R}_n$. These geometric parameters are shared across contrasts, whereas the amplitude $a_{n,e}$ and phase $\phi_{n,e}$ remain contrast specific. This factorized representation jointly leverages all acquired contrasts within a unified model, reducing redundant geometric parameterization and improving memory efficiency. The shared geometry captures common anatomical structure, while contrast-specific coefficients preserve amplitude and phase variations.

\subsection{Implicit Amplitude and Explicit Phase Modeling}
SGAM represents contrast-specific Gaussian amplitudes using a shared coordinate-conditioned INR. As illustrated in Fig.~\ref{fig:joint_reconstruction}, each Gaussian center $\boldsymbol{\mu}_n$ is first mapped to the unit cube using a fixed normalization $\nu(\cdot)$. A trainable multi-resolution hash encoding $h_{\theta_h}$~\cite{muller2022instant}, comprising $L$ levels with $F$ features per level, extracts spatial features from the normalized coordinate. A shared multi-output multilayer perceptron (MLP) $f_{\theta_f}$ then predicts the log-amplitudes of all contrasts
\begin{equation}
 \begin{aligned}
 \mathbf z_n&=f_{\theta_f}\!\left(h_{\theta_h}
       (\nu(\boldsymbol{\mu}_n))\right),      \mathbf z_n\in\mathbb R^E,\\
 a_{n,e}&=\exp\!( \operatorname{clip}(z_{n,e};b_-,b_+) ),
 \end{aligned}
 \label{eq:direct_decoder}
\end{equation}
where $z_{n,e}$ is the $e$-th element of $\mathbf z_n$, and $b_-<b_+$ are fixed clipping bounds. Clipping the log-amplitude before exponentiation prevents extreme amplitude values during optimization. 

The hash encoding provides spatial features at multiple resolutions, while one decoder output channel is assigned to each acquired contrast. Loss contributions from all contrasts update the same encoding and hidden layers. The resulting shared feature representation therefore couples amplitude estimation across both space and contrasts, while the separate output channels preserve contrast-dependent amplitudes. The encoding resolution and decoder capacity control the flexibility of this structured appearance model.

The phase of each Gaussian primitive and contrast is parameterized by an explicitly optimized variable $\eta_{n,e}\in\mathbb R$ and mapped to $(-\pi,\pi)$ as
\begin{equation}
 \phi_{n,e}=2\pi\operatorname{sigmoid}(\eta_{n,e})-\pi.
 \label{eq:phase_parameterization}
\end{equation}

This asymmetric parameterization combines spatially structured amplitude modeling and feature sharing across contrasts while retaining flexible contrast-specific phase variation. Together, the implicit amplitudes and explicit phases form the complex appearance coefficient $a_{n,e}\exp(\mathrm{i}\phi_{n,e})$ in Eq.~\eqref{eq:direct_image}.

\subsection{Scan-Specific Reconstruction}
\subsubsection{Initialization of 3D Gaussian Representation}
Gaussian centers are placed on a regularly subsampled image grid with stride $q$ along each spatial dimension. For a grid of size $D_1\times D_2\times D_3$, this gives $N=\prod_{d=1}^{3}\lceil D_d/q\rceil$ primitives, where $\lceil\cdot\rceil$ denotes the ceiling function. The initial rotations are set to the identity matrix, and the initial scales are determined from the distances between neighboring centers. 

For each contrast, let $x_e^{(0)}$ denote the zero-filled reconstruction. 
The phase of primitive $n$ is initialized as $\phi_{n,e} = \arg x_e^{(0)}(\boldsymbol{\mu}_n)$, and the corresponding variable $\eta_{n,e}$ is obtained using the inverse mapping defined by Eq.~\eqref{eq:phase_parameterization}.

The amplitude network is then prefitted to zero-filled log-magnitudes. The target for primitive $n$ and contrast $e$ is
\begin{equation}
 t_{n,e}=\operatorname{clip}\!(
 \log\!\left(|x_e^{(0)}(\boldsymbol{\mu}_n)|+\varepsilon\right);b_-,b_+),
\end{equation}
where $\varepsilon>0$ ensures numerical stability. For a mini-batch $\mathcal B$ of Gaussian centers, the prefitting loss is
\begin{equation}
 \mathcal{L}_{\mathrm{init}}=
 \frac{1}{|\mathcal B|E}\sum_{n\in\mathcal B}\sum_{e=1}^{E}
 (z_{n,e}-t_{n,e})^2.
 \label{eq:init_loss}
\end{equation}
This prefitting provides a data-dependent initialization of the implicit amplitude representation before joint reconstruction.

\subsubsection{Joint Optimization}
SGAM jointly optimizes the shared geometry, implicit amplitude model, and contrast-specific phases from the acquired k-space. The complete trainable parameter set is
\begin{equation}
 \begin{aligned}
 \Theta=\bigl\{\theta_h,\theta_f,
 \{\boldsymbol{\mu}_n,\boldsymbol{\rho}_n,\mathbf u_n\}_{n=1}^{N},
 \{\eta_{n,e}\}_{n=1,e=1}^{N,E}\bigr\}.
 \end{aligned}
\end{equation}
Here, $\boldsymbol{\rho}_n\in\mathbb R^3$ contains the log-scale parameters, such that $\boldsymbol{\sigma}_n=\exp(\boldsymbol{\rho}_n)$ elementwise, and $\mathbf u_n\in\mathbb R^4$ is a trainable quaternion. The rotation matrix $\boldsymbol{R}_n$ is constructed from the normalized quaternion $\mathbf u_n/\|\mathbf u_n\|_2$. 

The joint reconstruction objective consists of multi-coil k-space data consistency and magnitude total-variation regularization
\begin{equation}
 \mathcal{L}(\Theta)=\mathcal{L}_{\mathrm{dc}}
 (\Theta)+\lambda_{\mathrm{tv}}\mathcal{L}_{\mathrm{tv}}(\Theta),
 \label{eq:direct_loss}
\end{equation}
where $\lambda_{\mathrm{tv}}\geq 0$ controls the regularization strength. The data-consistency term $\mathcal{L}_{\mathrm{dc}}$ is the $L_1$ loss between predicted and measured k-space samples across all receiver coils and contrasts. The term $\mathcal{L}_{\mathrm{tv}}$ applies 3D total variation~\cite{block2007undersampled} to the reconstructed magnitudes $m_e(\mathbf r)=|\widehat{x}_e(\mathbf r)|$ to promote spatial coherence. The overall loss is evaluated across all contrasts, allowing each contrast to contribute to updates of the shared parameters.

Gradients from the amplitude network are stopped at its coordinate inputs. Consequently, Gaussian centers are updated only through differentiable rasterization rather than through the coordinate-conditioned amplitude pathway. Their updated locations are used for the subsequent network query, and the total number of primitives remains fixed throughout optimization.

After optimization, the predicted k-space data are replaced with the measured data at acquired locations. The final image is then obtained by inverse Fourier transformation and sensitivity-weighted coil combination.

\section{Experiments}
\label{sec:experiments}
\subsection{Datasets and Sampling Schemes}
\label{sec:datasets_sampling}
This study was approved by the Institutional Review Board of the Shenzhen Institutes of Advanced Technology, Chinese Academy of Sciences.

Two types of 3D multi-contrast MRI data were included. Fully sampled MULTIPLEX (MTP) data~\cite{ye2022multi} were acquired from three participants on a 3~T scanner (uMR 790, United Imaging Healthcare, Shanghai). Each MTP scan comprised two six-contrast groups acquired at flip angles of $4^{\circ}$ and $16^{\circ}$, which were reconstructed independently. Fully sampled two-point DIXON CMRA data were acquired from three participants on a 5~T scanner (United Imaging Healthcare, Shanghai). The detailed imaging parameters are listed in Table~\ref{tab:datasets}.
\begin{table}[t]
\centering
\caption{Acquisition and reconstruction characteristics.}
\label{tab:datasets}
\footnotesize
\setlength{\tabcolsep}{3pt}
\renewcommand{\arraystretch}{1.15}
\begin{tabular}{lcc}
\hline
Characteristic & Retrospective MTP & Retrospective CMRA  \\
\hline
Field strength & 3~T & 5~T  \\
Volumes per scan & 12 ($2\times6$) & 2  \\
voxel size (mm$^3$) & $0.68\times0.68\times2.0$ & $1.2\times1.2\times1.5$  \\
matrix & $336\times288\times84$ & $240\times\{400,374,424\}\times92$ \\
Original $\rightarrow$ virtual coils & $32\rightarrow12$ & $\{48,32,48\}\rightarrow16$ \\
\hline
\end{tabular}
\end{table}

In all experiments, undersampling was applied along the phase- and partition-encoding dimensions, while the readout dimension remained fully sampled. Variable-density Poisson-disc masks~\cite{vasanawala2011practical} were generated using SigPy~\cite{ong2019sigpy} at acceleration factors of $12\times$ and $16\times$. The two CMRA contrasts used different masks, whereas six contrast-specific masks were generated for MTP and reused across the two flip-angle groups. Images reconstructed from the fully sampled data served as references.

BART~\cite{blumenthal2022bart} was used for SVD-based coil compression~\cite{buehrer2007array} and ESPIRiT coil-sensitivity estimation~\cite{uecker2014espirit}. For each scan, the same coil-compression matrix and sensitivity maps were used across all contrasts.

\subsection{Comparison Methods and Implementation Details}
\label{sec:comparison_evaluation}
\textit{Comparison methods:}
SGAM was compared with five representative scan-specific reconstruction methods, including ZS-SSL~\cite{yaman2022zero}, ConvDecoder~\cite{darestani2021accelerated}, LRTFR~\cite{luo2023low}, 2DINR~\cite{niessen2025inr}, and PICS~\cite{lustig2007sparse}. ZS-SSL~\cite{yaman2022zero} is a zero-shot self-supervised method and was adapted to a 3D unrolled architecture for joint multi-contrast reconstruction. ConvDecoder~\cite{darestani2021accelerated} is a deep-image-prior method and was implemented as a 3D convolutional generator with multi-contrast outputs. LRTFR~\cite{luo2023low} is a functional tensor representation; our implementation jointly modeled 3D multi-contrast volumes using a learnable core tensor and coordinate-dependent SIREN factors~\cite{sitzmann2020implicit}, without rank reduction. 2DINR~\cite{niessen2025inr} is a 2D coordinate-based implicit neural representation using hash encoding. After an inverse Fourier transform along the fully sampled readout direction, an independent INR was optimized for each resulting 2D slice to jointly reconstruct all contrasts. PICS~\cite{lustig2007sparse} is a conventional compressed-sensing method and was implemented in BART to reconstruct each 3D contrast independently using wavelet and TV regularization.

\textit{Implementation details:}
For SGAM, the multi-resolution hash encoding used 16 levels with two features per level, a base resolution of 16, a per-level scale factor of 1.34, and at most $2^{20}$ entries per level. Gaussian centers were initialized with a grid stride of three. The amplitude MLP had two hidden layers with 64 units for two-contrast CMRA and 256 units for six-contrast MTP. The TV coefficient was $5\times10^{-4}$ for MTP and $10^{-3}$ for CMRA. SGAM was optimized using Adam with parameter-specific learning rates and implemented in PyTorch using tiny-cuda-nn for the hash encoding and a CUDA-based complex Gaussian rasterizer. For the comparison methods, method-specific configurations, including regularization weights and optimization settings, were configured separately for the MTP and CMRA tasks. All experiments were conducted on NVIDIA A100 80-GB PCIe GPUs.

\textit{Evaluation metrics:}
Reconstruction quality was evaluated against the fully sampled reference images using peak signal-to-noise ratio (PSNR), structural similarity index (SSIM), and normalized root-mean-square error (NRMSE). Higher PSNR and SSIM and lower NRMSE indicate better agreement with the reference.

\subsection{Ablation Study Design}
\label{sec:ablation_design}
Ablation experiments were conducted on the undersampled MTP and CMRA datasets at $16\times$ to evaluate the contributions of shared Gaussian geometry, implicit amplitude modeling, and TV regularization. Five configurations were evaluated. Joint configurations reconstructed each six-contrast MTP group or the two CMRA contrasts together, whereas independent configurations reconstructed each contrast separately. Phase was explicitly optimized in all configurations; 'implicit' and 'explicit' refer only to the amplitude parameterization.
\begin{enumerate}
\item \textit{Independent explicit:} Each contrast was reconstructed separately using independent Gaussian geometry and directly optimized amplitude coefficients.
\item \textit{Joint explicit:} All contrasts within a reconstruction group shared the Gaussian geometry, while their contrast-specific amplitude coefficients were directly optimized.
\item \textit{Independent implicit:} Each contrast was reconstructed using independent Gaussian geometry and a separate single-output amplitude network.
\item \textit{Joint implicit (SGAM):} All contrasts within a reconstruction group shared the Gaussian geometry and were jointly modeled by a single multi-output amplitude network.
\item \textit{SGAM without TV:} This configuration was identical to SGAM except that the TV regularization coefficient was set to zero.
\end{enumerate}

\section{Results}
\subsection{Reconstruction Results}
\label{sec:Retrospective Reconstruction Results}
\begin{table*}[t]
\centering
\caption{Retrospective reconstruction results for the MTP and CMRA tasks at acceleration factors of $12\times$ and $16\times$. Metrics are reported as the mean and standard deviation across subjects. The best and second-best mean values are shown in bold and underlined, respectively.}
\label{tab:retrospective_results}
\footnotesize
\setlength{\tabcolsep}{3pt}
\renewcommand{\arraystretch}{1.10}
\resizebox{\textwidth}{!}{%
\begin{tabular}{@{}cclcccccc@{}}
\hline
Dataset & Accel. & Metric & SGAM & ZS-SSL & ConvDecoder & LRTFR & 2DINR & PICS \\
\hline
\multirow{6}{*}{MTP}
 & \multirow{3}{*}{$12\times$}
 & PSNR (dB)$\uparrow$ & $\mathbf{37.08 \pm 1.00}$ & \underline{$36.66 \pm 1.04$} & $35.74 \pm 1.06$ & $34.31 \pm 1.12$ & $34.84 \pm 1.49$ & $34.03 \pm 1.36$ \\
 & & SSIM$\uparrow$ & $\mathbf{0.9341 \pm 0.0099}$ & $0.9069 \pm 0.0103$ & \underline{$0.9071 \pm 0.0158$} & $0.8827 \pm 0.0118$ & $0.8617 \pm 0.0218$ & $0.8537 \pm 0.0234$ \\
 & & NRMSE$\downarrow$ & $\mathbf{0.0143 \pm 0.0016}$ & \underline{$0.0150 \pm 0.0017$} & $0.0167 \pm 0.0020$ & $0.0197 \pm 0.0025$ & $0.0189 \pm 0.0033$ & $0.0204 \pm 0.0031$ \\
\cline{2-9}
 & \multirow{3}{*}{$16\times$} & PSNR (dB)$\uparrow$ & $\mathbf{36.39 \pm 1.02}$ & \underline{$35.27 \pm 1.09$} & $35.13 \pm 1.07$ & $33.46 \pm 1.13$ & $33.52 \pm 1.52$ & $32.81 \pm 1.29$ \\
 & & SSIM$\uparrow$ & $\mathbf{0.9262 \pm 0.0118}$ & $0.8873 \pm 0.0142$ & \underline{$0.8971 \pm 0.0162$} & $0.8620 \pm 0.0127$ & $0.8350 \pm 0.0265$ & $0.8480 \pm 0.0222$ \\
 & & NRMSE$\downarrow$ & $\mathbf{0.0155 \pm 0.0018}$ & \underline{$0.0176 \pm 0.0021$} & $0.0179 \pm 0.0022$ & $0.0217 \pm 0.0028$ & $0.0220 \pm 0.0040$ & $0.0235 \pm 0.0034$ \\
\hline
\multirow{6}{*}{CMRA} & \multirow{3}{*}{$12\times$} & PSNR (dB)$\uparrow$ & $\mathbf{42.42 \pm 3.62}$ & \underline{$41.97 \pm 3.58$} & $40.97 \pm 3.74$ & $40.55 \pm 3.24$ & $40.03 \pm 3.70$ & $40.91 \pm 3.85$ \\
 & & SSIM$\uparrow$ & $\mathbf{0.9619 \pm 0.0160}$ & \underline{$0.9598 \pm 0.0179$} & $0.9504 \pm 0.0196$ & $0.9473 \pm 0.0155$ & $0.9436 \pm 0.0190$ & $0.9494 \pm 0.0202$ \\
 & & NRMSE$\downarrow$ & $\mathbf{0.0080 \pm 0.0032}$ & \underline{$0.0084 \pm 0.0034$} & $0.0095 \pm 0.0040$ & $0.0098 \pm 0.0034$ & $0.0106 \pm 0.0043$ & $0.0096 \pm 0.0043$ \\
\cline{2-9}
 & \multirow{3}{*}{$16\times$} & PSNR (dB)$\uparrow$ & $\mathbf{41.77 \pm 3.64}$ & \underline{$41.21 \pm 3.53$} & $40.03 \pm 3.87$ & $39.80 \pm 3.19$ & $39.16 \pm 3.78$ & $39.68 \pm 3.90$ \\
 & & SSIM$\uparrow$ & $\mathbf{0.9568 \pm 0.0171}$ & \underline{$0.9543 \pm 0.0194$} & $0.9418 \pm 0.0212$ & $0.9393 \pm 0.0160$ & $0.9345 \pm 0.0202$ & $0.9370 \pm 0.0219$ \\
 & & NRMSE$\downarrow$ & $\mathbf{0.0087 \pm 0.0035}$ & \underline{$0.0092 \pm 0.0036$} & $0.0106 \pm 0.0045$ & $0.0107 \pm 0.0037$ & $0.0117 \pm 0.0048$ & $0.0111 \pm 0.0048$ \\
\hline
\end{tabular}%
}
\end{table*}
Table~\ref{tab:retrospective_results} summarizes the results for the MTP and CMRA tasks. SGAM consistently achieved the best performance across both tasks and acceleration factors. Its SSIM advantage was particularly pronounced for MTP and became more evident at $16\times$, demonstrating robust structural preservation under severe undersampling.

\begin{figure*}[t]
\centering
\includegraphics[width=\textwidth]{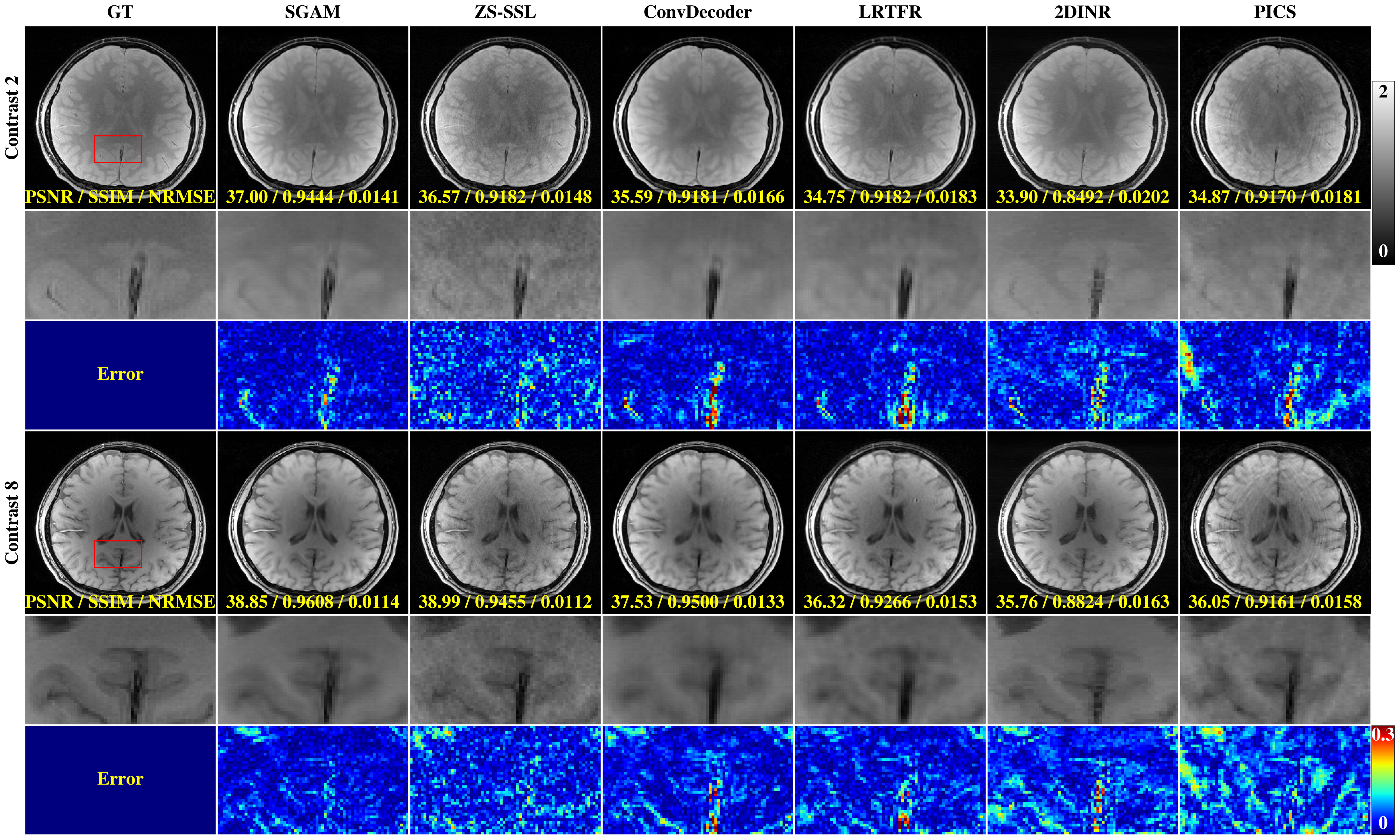}
\caption{Representative retrospective MTP reconstructions for one subject at $12\times$ acceleration. Contrasts 2 and 8 are selected from the first and second independently reconstructed six-contrast groups, respectively. The overlaid values report whole-volume PSNR (dB), SSIM, and NRMSE. For each contrast, the rows show reconstructed magnitude images, enlarged views of the region marked in the reference, and absolute-error maps.}
\label{fig:retrospective_mtp_r12}
\end{figure*}

\begin{figure*}[t]
\centering
\includegraphics[width=\textwidth]{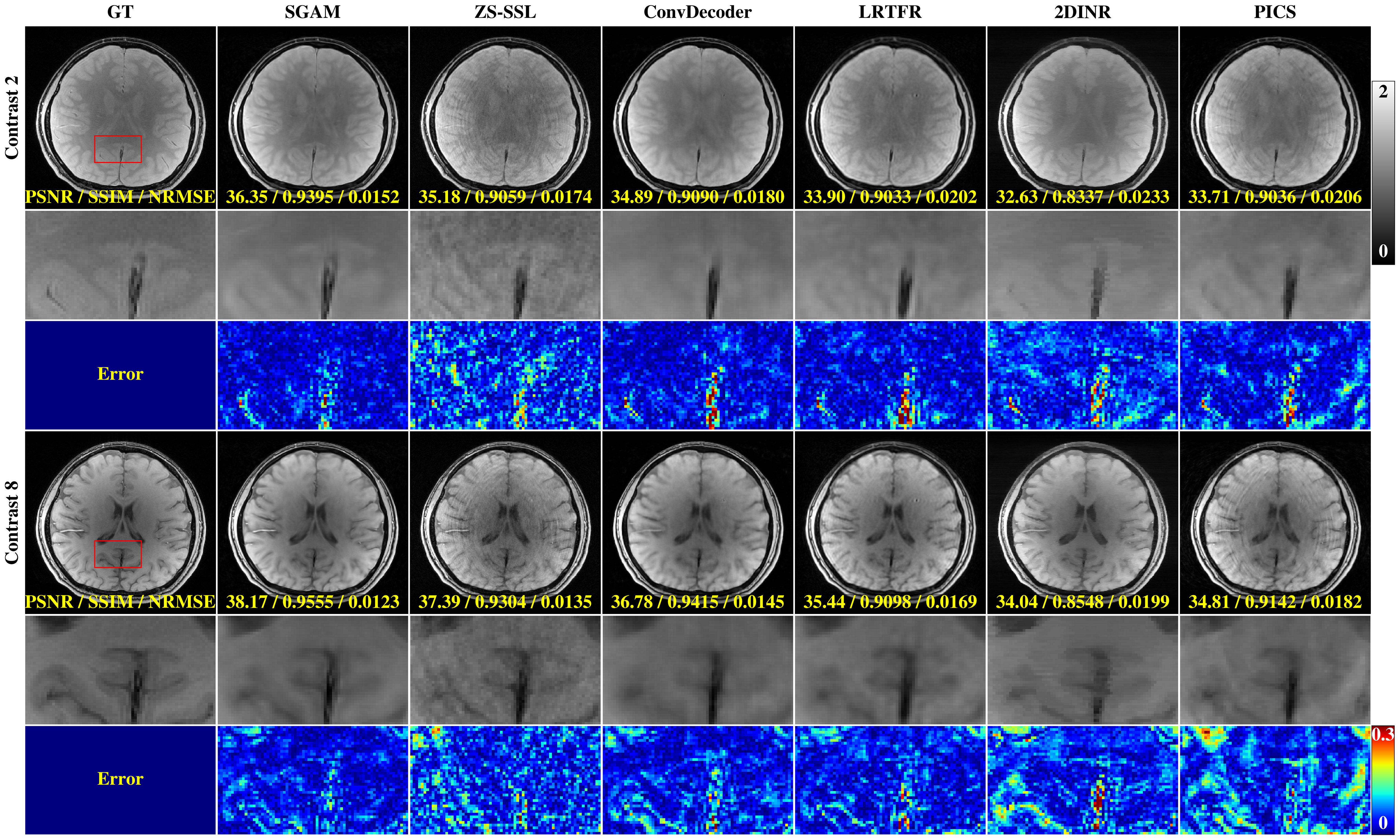}
\caption{Representative retrospective MTP reconstructions for one subject at $16\times$ acceleration. Contrasts 2 and 8 are selected from the first and second independently reconstructed six-contrast groups, respectively. The overlaid values report whole-volume PSNR (dB), SSIM, and NRMSE. For each contrast, the rows show reconstructed magnitude images, enlarged views of the region marked in the reference, and absolute-error maps.}
\label{fig:retrospective_mtp_r16}
\end{figure*}

\begin{figure*}[t]
\centering
\includegraphics[width=\textwidth]{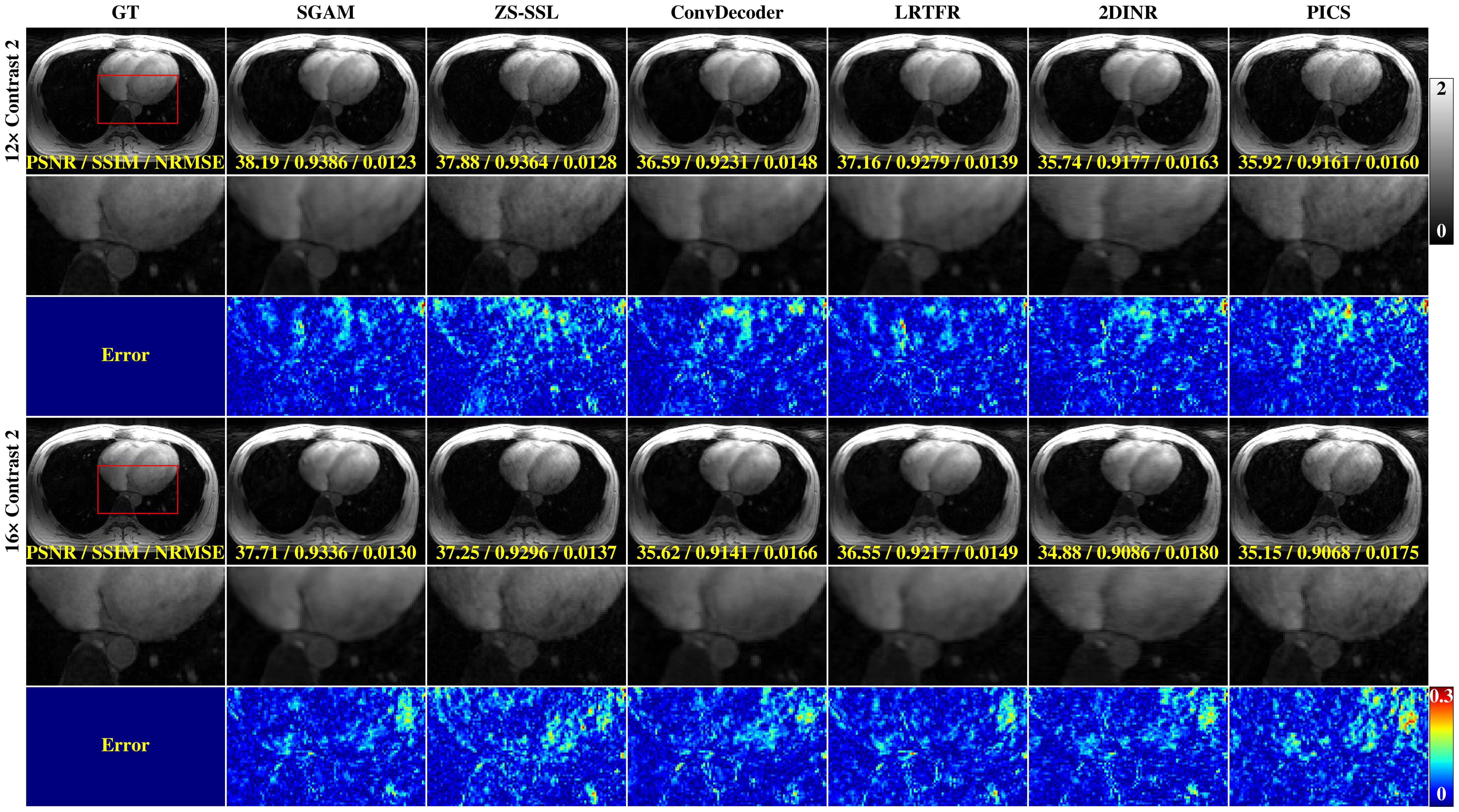}
\caption{Representative retrospective CMRA reconstructions for one subject at $12\times$ and $16\times$ acceleration. Contrast 2 is shown. The overlaid values report whole-volume PSNR (dB), SSIM, and NRMSE. For each acceleration factor, the rows show reconstructed magnitude images, enlarged views of the region marked in the reference, and absolute-error maps.}
\label{fig:retrospective_cmra_r16}
\end{figure*}
Figs.~\ref{fig:retrospective_mtp_r12} and~\ref{fig:retrospective_mtp_r16} show the MTP reconstruction results at $12\times$ and $16\times$ acceleration, respectively, while Fig.~\ref{fig:retrospective_cmra_r16} presents the CMRA results at both acceleration factors. The visual comparisons show consistent trends across the two tasks. SGAM better preserves small vessels, tissue boundaries, and local textures, with improved structural continuity and fewer residual errors in the enlarged regions. ZS-SSL retains the principal anatomical structures but exhibits residual aliasing and spatial noise, leading to errors distributed over broader regions. ConvDecoder and LRTFR tend to produce smoother reconstructions, resulting in blurred boundaries and the loss of fine structures and high-frequency details. 2DINR shows local structural degradation, spatially nonuniform artifacts, and inter-slice inconsistency, whereas PICS suffers from noise amplification and structured residual artifacts. As the acceleration factor increases from $12\times$ to $16\times$, these degradations become more pronounced for all methods, while SGAM maintains comparatively stable image quality and detail preservation.

\subsection{Quantitative Mapping Results}
\begin{figure*}[t]
\centering
\includegraphics[width=\textwidth]{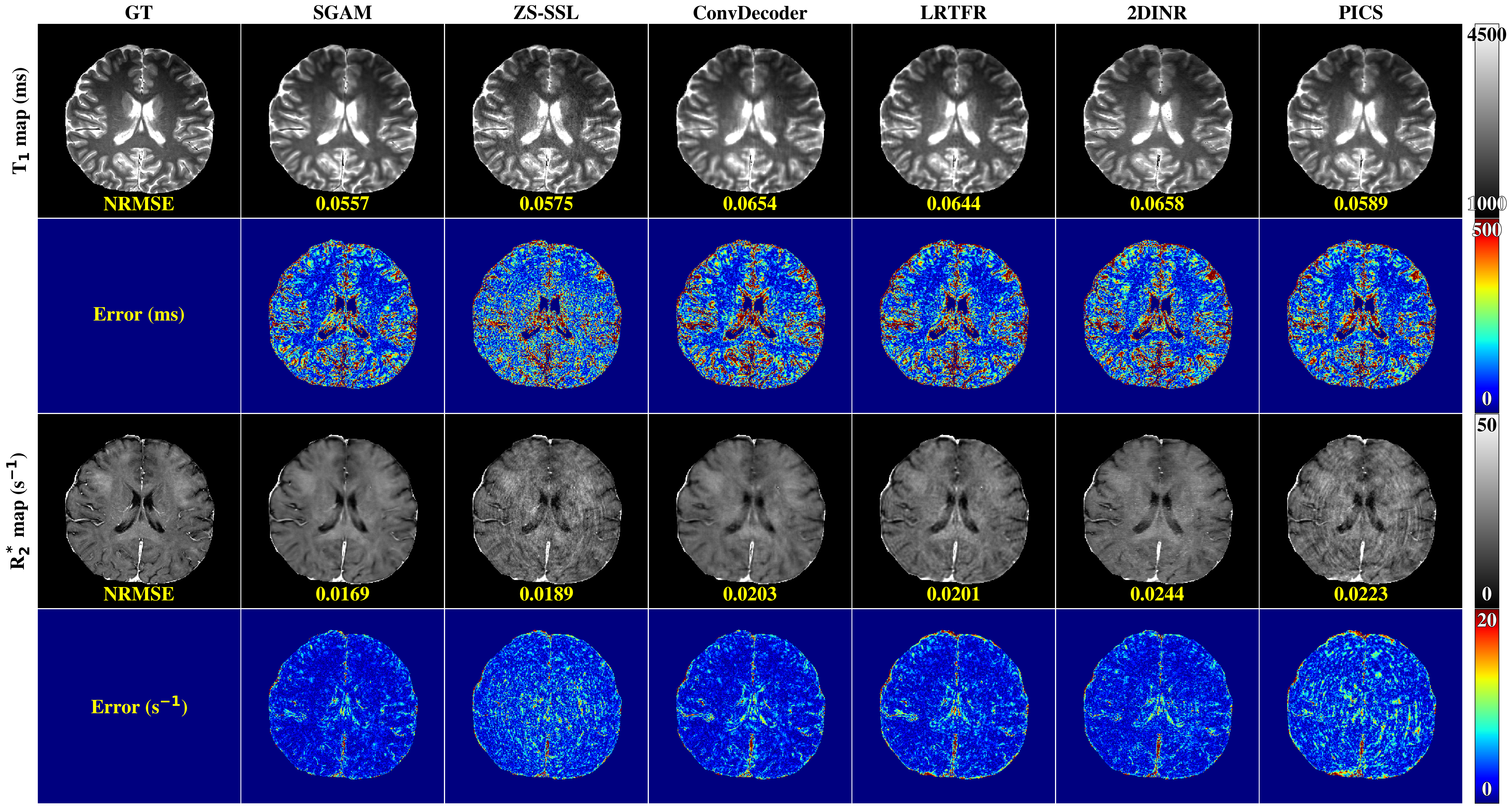}
\caption{Quantitative mapping results for a representative MTP subject retrospectively undersampled at $16\times$. The overlaid values report NRMSE calculated within the displayed slice. The first and third rows show the derived $T_1$ maps and R$_2^*$ maps, respectively, while the second and fourth rows show their corresponding absolute-error maps relative to GT.}
\label{fig:mtp_mapping_r16}
\end{figure*}
Fig.~\ref{fig:mtp_mapping_r16} compares the quantitative mapping results at $16\times$ acceleration. The $T_1$ and $R_2^*$ maps were fitted from different subsets of the 12 reconstructed MTP contrast images. SGAM achieved the lowest NRMSE for both $T_1$ and $R_2^*$ and better preserved tissue structure and ventricular boundaries, with fewer errors around tissue interfaces. ZS-SSL retained the main anatomical structures but exhibited more widespread residual errors in both maps. ConvDecoder produced overly smooth maps with blurred local variations, whereas LRTFR and 2DINR showed larger localized deviations around tissue boundaries. PICS recovered a relatively accurate $T_1$ map but showed more spatially distributed errors in the $R_2^*$ map. Quantitative mapping accuracy depends on relative signal relationships across acquisition settings or echo times rather than solely on the reconstruction accuracy of individual contrasts. Correlated errors may partially cancel, whereas acquisition- or echo-dependent distortions can be amplified during nonlinear fitting. These results indicate that SGAM more consistently preserves both anatomical structure and the inter-volume signal relationships required for downstream quantitative mapping.

\subsection{Ablation Results}
\label{sec:ablation_results}
\begin{table*}[t]
\centering
\caption{Ablation results for the MTP and CMRA tasks at retrospective acceleration factors of $16\times$. Metrics are reported as the mean and standard deviation across subjects. The best and second-best mean values are shown in bold and underlined, respectively.}
\label{tab:ablation_results}
\small
\setlength{\tabcolsep}{3pt}
\renewcommand{\arraystretch}{1.15}
\begin{tabular*}{\textwidth}{@{\extracolsep{\fill}}clccccc@{}}
\hline
Dataset & Metric
 & \shortstack{Independent\\explicit}
 & \shortstack{Joint\\explicit}
 & \shortstack{Independent\\implicit}
 & \shortstack{Joint implicit\\(SGAM)}
 & \shortstack{SGAM\\without TV} \\
\hline
\multirow{3}{*}{MTP}
  & PSNR (dB)$\uparrow$ & $35.91 \pm 1.11$ & $36.27 \pm 1.05$ & $\underline{36.39 \pm 1.06}$ & $\mathbf{36.39 \pm 1.02}$ & $36.09 \pm 1.04$ \\
 & SSIM$\uparrow$ & $0.9172 \pm 0.0150$ & $0.9229 \pm 0.0130$ & $\mathbf{0.9323 \pm 0.0114}$ & $\underline{0.9262 \pm 0.0118}$ & $0.9166 \pm 0.0149$ \\
 & NRMSE$\downarrow$ & $0.0164 \pm 0.0020$ & $0.0157 \pm 0.0018$ & $\underline{0.0155 \pm 0.0019}$
 & $\mathbf{0.0155 \pm 0.0018}$ & $0.0160 \pm 0.0019$ \\
\hline
\multirow{3}{*}{CMRA}
 & PSNR (dB)$\uparrow$ & $40.73 \pm 3.78$ & $40.89 \pm 3.77$ & $\underline{41.34 \pm 3.82}$ & $\mathbf{41.77 \pm 3.64}$ & $41.30 \pm 3.63$ \\
 & SSIM$\uparrow$ & $0.9451 \pm 0.0196$ & $0.9471 \pm 0.0192$ & $\underline{0.9530 \pm 0.0187}$ & $\mathbf{0.9568 \pm 0.0171}$ & $0.9520 \pm 0.0180$ \\
 & NRMSE$\downarrow$ & $0.0098 \pm 0.0041$ & $0.0096 \pm 0.0040$ & $0.0091 \pm 0.0039$ & $\mathbf{0.0087 \pm 0.0035}$ & $\underline{0.0091 \pm 0.0037}$ \\
\hline
\end{tabular*}
\end{table*}
Table~\ref{tab:ablation_results} evaluates geometric sharing, implicit amplitude modeling, and TV regularization at $16\times$ acceleration. Joint explicit reconstruction consistently outperformed independent explicit reconstruction across both tasks, demonstrating the benefit of representing common anatomy with shared Gaussian geometry. In both the joint and independent settings, implicit amplitude modeling also consistently improved over the corresponding explicit representation, supporting its role as a structured appearance prior.

The effect of joint implicit reconstruction was task dependent. For six-contrast MTP, SGAM achieved PSNR and NRMSE comparable to independent implicit reconstruction, although the latter yielded higher SSIM, possibly due to its greater contrast-specific representation capacity. In contrast, for two-contrast CMRA, SGAM achieved the best performance across all metrics. For complete-dataset reconstruction, joint implicit reconstruction was also more time-efficient, reducing the average total runtime from 5299.7 to 2702.1~s for MTP and from 1509.4 to 1121.4~s for CMRA. These findings reveal a trade-off between contrast-specific representation capacity and the regularization provided by cross-contrast sharing, while demonstrating the runtime advantage of joint optimization.

Removing TV regularization degraded performance across both tasks, confirming that TV provides a complementary spatial constraint beyond shared Gaussian geometry and implicit amplitude modeling.

\section{Discussion}
\subsection{Reconstruction Performance Analysis}
SGAM jointly reconstructs 3D multi-contrast MRI within a unified compact volumetric representation. By sharing anatomical geometry while preserving contrast-specific appearance, SGAM reduces redundant parameterization and improves reconstruction accuracy. ZS-SSL and ConvDecoder were the strongest competing methods, supporting the importance of jointly exploiting 3D spatial structure and inter-contrast correlations. However, their dense 3D convolutional architectures imposed substantial computational costs and required reduced network capacity for full-volume reconstruction. The factorized representation of LRTFR provided limited capacity for complex contrast-dependent details, resulting in oversmoothing of fine structures. The slice-wise 2DINR did not enforce through-plane consistency, leading to inter-slice discontinuities and local artifacts. PICS reconstructed each contrast independently and could not exploit complementary cross-contrast information. Overall, these comparisons highlight the importance of combining full-volume 3D consistency with flexible cross-contrast information sharing.

\subsection{Computational Considerations}
\begin{table}[t]
\centering
\caption{Computational costs of the compared methods on MTP and CMRA, averaged over acceleration factors of $12\times$ and $16\times$.}
\label{tab:comparison_computational_cost}
\small
\setlength{\tabcolsep}{3pt}
\renewcommand{\arraystretch}{1.15}
\begin{tabular*}{\columnwidth}{@{\extracolsep{\fill}}clcc@{}}
\hline
Dataset & Method
 & \shortstack{Recorded runtime\\(s)$\downarrow$}
 & \shortstack{Peak allocated\\GPU memory (GiB)$\downarrow$} \\
\hline
\multirow{6}{*}{MTP}
 & SGAM (ours) & $2702.1 \pm 26.3$ & $33.95$ \\
 & ZS-SSL$^{\dagger}$ & $23171.9 \pm 69.9$ & $65.09$ \\
 & ConvDecoder$^{\dagger}$ & $16693.6 \pm 56.9$ & $38.51$ \\
 & LRTFR & $7268.6 \pm 17.5$ & $34.92$ \\
 & 2DINR & $1732.9 \pm 120.6$ & $0.16$ \\
\hline
\multirow{6}{*}{CMRA} 
 & SGAM (ours) & $1121.4 \pm 96.6$ & $19.15$ \\
 & ZS-SSL$^{\dagger}$ & $1825.5 \pm 173.0$ & $32.23$ \\
 & ConvDecoder$^{\dagger}$ & $6929.7 \pm 363.1$ & $27.48$ \\
 & LRTFR & $1891.5 \pm 140.1$ & $17.12$ \\
 & 2DINR & $670.8 \pm 31.7$ & $0.21$ \\
\hline
\end{tabular*}
\par\vspace{3pt}
\begin{minipage}{\columnwidth}
\footnotesize
Costs correspond to reconstructing a complete dataset: 12 MTP volumes or two CMRA contrasts. For methods processing contrasts or reconstruction groups sequentially, runtimes were summed and peak memory was taken as the maximum across runs. GPU memory denotes the peak memory allocated to PyTorch tensors ($1\,\mathrm{GiB}=2^{30}$ bytes). $^{\dagger}$ZS-SSL and ConvDecoder used automatic mixed precision (AMP). 
\end{minipage}
\end{table}
Under the evaluated implementations, SGAM achieved higher reconstruction accuracy with shorter recorded runtime than ZS-SSL, ConvDecoder, and LRTFR for both MTP and CMRA (Table~\ref{tab:comparison_computational_cost}). The runtime advantage was particularly pronounced for 12-volume MTP reconstruction, where shared geometry and joint amplitude modeling reduced repeated computation across contrasts. SGAM also required less peak GPU memory than the convolutional baselines and a comparable amount to LRTFR. Despite using mixed precision, ZS-SSL and ConvDecoder retained high memory demands due to volumetric multi-contrast feature processing. 2DINR was faster and substantially more memory-efficient through slice-wise processing, but produced lower reconstruction quality and lacked a unified 3D representation. Because PICS was executed on the CPU, its GPU efficiency was not directly comparable. These results demonstrate that SGAM provides a favorable balance between reconstruction quality and computational cost among the evaluated joint 3D multi-contrast methods.

\subsection{Effect of Gaussian Initialization Density}
\begin{table*}[t]
\centering
\caption{Effect of Gaussian initialization density on reconstruction performance and computational cost at $16\times$ acceleration. Results are shown for one six-contrast MTP group and one two-contrast CMRA subject. The best value for each metric within each dataset is shown in bold.}
\label{tab:initialization_density}
\small
\setlength{\tabcolsep}{4pt}
\renewcommand{\arraystretch}{1.12}
\begin{tabular*}{\textwidth}{@{\extracolsep{\fill}}cccccccc@{}}
\hline
Dataset & Step & \shortstack{Gaussian\\count}
 & \shortstack{PSNR\\(dB)$\uparrow$}
 & SSIM$\uparrow$ & NRMSE$\downarrow$
 & \shortstack{Total time\\(s)$\downarrow$}
 & \shortstack{Peak memory\\(GiB)$\downarrow$} \\
\hline
\multirow{3}{*}{MTP}
 & $3^{\dagger}$ & 301,056 & $\mathbf{36.21}$ & $\mathbf{0.9254}$ & $\mathbf{0.0155}$ & 1378.7 & 33.95 \\
 & 4 & 127,008 & 35.24 & 0.9012 & 0.0173 & 1250.3 & 33.90 \\
 & 5 & 67,048 & 34.21 & 0.8828 & 0.0195 & $\mathbf{1175.8}$ & $\mathbf{33.89}$ \\
\hline
\multirow{3}{*}{CMRA}
 & $3^{\dagger}$ & 332,320 & $\mathbf{38.23}$ & $\mathbf{0.9397}$ & $\mathbf{0.0123}$ & 1221.8 & 19.18 \\
 & 4 & 138,000 & 38.10 & 0.9389 & 0.0125 & 731.5 & 19.14 \\
 & 5 & 72,960 & 37.92 & 0.9364 & 0.0127 & $\mathbf{517.8}$ & $\mathbf{19.13}$ \\
\hline
\end{tabular*}
\par\vspace{3pt}
\begin{minipage}{\textwidth}
\footnotesize
$\dagger$ Setting used in the main experiments. Step denotes the subsampling interval of the initial 3D grid; a larger step produces fewer Gaussian primitives.
\end{minipage}
\end{table*}
Table~\ref{tab:initialization_density} characterizes the trade-off among Gaussian initialization density, reconstruction accuracy, and computational cost. Increasing the grid step from 3 to 5 reduced the number of Gaussian primitives by approximately $78\%$ for both tasks, resulting in lower runtime but monotonically degraded reconstruction quality. This trade-off was task dependent: MTP showed a pronounced accuracy reduction with only a $15\%$ runtime saving, indicating that its richer fine-scale structures require denser Gaussian support, whereas CMRA exhibited only minor accuracy degradation with a $58\%$ runtime saving and was therefore more tolerant of sparse initialization. Peak memory remained nearly unchanged, suggesting that it was dominated by full-volume image, k-space, and encoding operations rather than Gaussian parameters. Grid step 3 was therefore retained to prioritize reconstruction accuracy, particularly for MTP, while further memory reductions would require more efficient forward encoding or chunked coil and contrast processing rather than simply reducing the number of Gaussian primitives.

\subsection{Limitations and Future Work}
Nevertheless, this study has several limitations. First, for multi-contrast data with complex spatial textures, the current joint amplitude model may provide limited contrast-specific representation capacity. Future work will explore more expressive amplitude models that better capture contrast-dependent variations while maintaining cross-contrast information sharing. Second, despite the compact Gaussian parameterization, full-volume encoding and data-consistency operations still require substantial GPU memory. Chunked processing across coils and contrasts, blockwise or multi-resolution optimization, and memory-efficient data-consistency schemes may improve scalability.

\section{Conclusion}
We proposed SGAM, a memory-efficient scan-specific framework for 3D MCMRI reconstruction. SGAM represents common anatomical structure using shared anisotropic Gaussian geometry, while implicitly modeled amplitudes and explicitly optimized phases preserve contrast-specific complex appearance. Experiments demonstrated consistent improvements over the comparison methods across imaging tasks and acceleration factors, while ablation studies confirmed the complementary contributions of geometry sharing, implicit amplitude modeling, and spatial regularization. SGAM also achieved a favorable balance between reconstruction quality and computational cost. These findings demonstrate the effectiveness of shared Gaussian geometry and implicit contrast-dependent amplitude modeling for 3D MCMRI reconstruction.


\bibliographystyle{IEEEtran}
\section*{References}
\bibliography{gaussian_ref}

\end{document}